\documentclass[runningheads]{llncs}
\usepackage[T1]{fontenc}
\usepackage{graphicx}
\usepackage{booktabs}
\usepackage{float}
\usepackage{pifont}
\usepackage{makecell}
\usepackage{amsmath}
\usepackage{caption}
\usepackage{subcaption}
\newcommand{\cmark}{\ding{51}} 
\newcommand{\xmark}{\ding{55}} 

\begin{document}
\title{CzechTopic: A Benchmark for Zero-Shot Topic Localization in Historical Czech Documents}
\titlerunning{CzechTopic}
%
\author{Martin Kostelník\inst{1}\orcidID{0009-0002-5478-9580} \and
Michal Hradiš\inst{1}\orcidID{0000-0002-6364-129X} \and Martin Dočekal\inst{1}\orcidID{0000-0002-0580-9357}}

%
\authorrunning{M. Kostelník et al.}
%
\institute{Faculty of Information Technology, Brno University of Technology,\\Brno, Czech Republic\\ \email{\{ikostelnik,ihradis,idocekal\}@fit.vut.cz}}
%
\maketitle              
\begin{abstract}
Topic localization aims to identify spans of text that express a given topic defined by a name and description. To study this task, we introduce a human-annotated benchmark based on Czech historical documents, containing human-defined topics together with manually annotated spans and supporting evaluation at both document and word levels. Evaluation is performed relative to human agreement rather than a single reference annotation. We evaluate a diverse range of large language models alongside BERT-based models fine-tuned on a distilled development dataset. Results reveal substantial variability among LLMs, with performance ranging from near-human topic detection to pronounced failures in span localization. While the strongest models approach human agreement, the distilled token embedding models remain competitive despite their smaller scale. The dataset and evaluation framework are publicly available at: \url{https://github.com/dcgm/czechtopic}.

\end{abstract}
\section{Introduction}
\begin{figure}[ht]
\centering
\includegraphics[width=\textwidth]{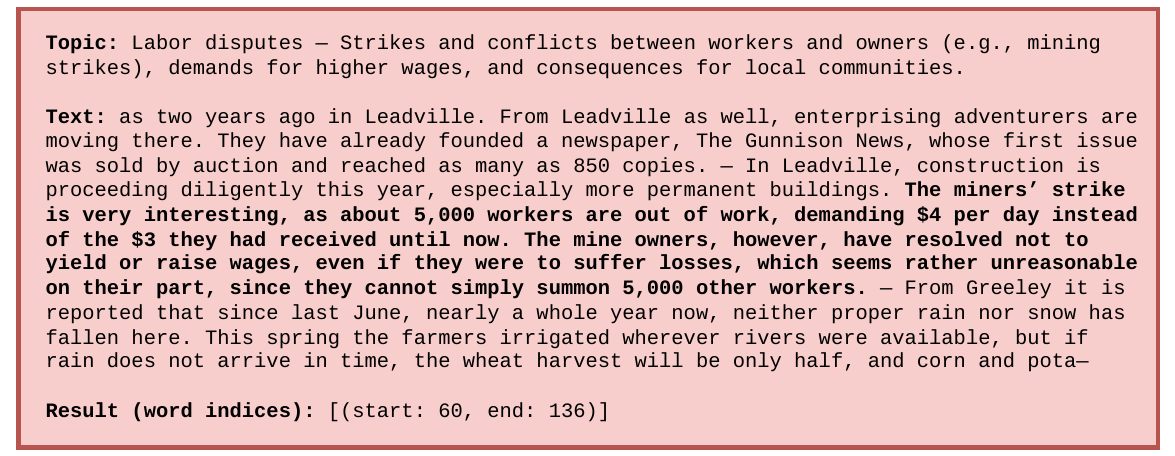}
\caption{Illustration of the topic localization task. A topic is defined by a name and a textual description. The objective is to identify and localize all spans in the document that correspond to the given topic.}
\label{fig:task-showcase}
\end{figure}

Topic localization is the task of identifying the exact spans of text that express a given topic. It is a largely underexplored problem in document understanding. Previous work has extensively explored document-level classification, topic discovery, and topic segmentation \cite{bert,adhikari2019docbert,grootendorst2022bertopic,wu2024fastopic,wikisection}. These approaches focus on identifying document themes or detecting topic boundaries, rather than localizing where a topic is expressed within the text itself. Compared to document classification or sentence-level labeling, topic localization requires word-level boundary decisions, allows overlapping and non-exclusive spans, and permits multiple disjoint mentions of the same topic (Figure~\ref{fig:task-showcase}).

Topic localization is motivated by applications where analysts need to understand not only whether a topic appears in a document, but also where it is discussed. Such fine-grained annotations support both qualitative and quantitative analysis of text collections, particularly in digital humanities and historical research. For example, historians may trace social or political themes across archival materials. Compared to document-level classification, span-level localization enables richer downstream uses such as automatic or semi-automatic tagging, evidence extraction, topic-based navigation, and assisted corpus annotation, allowing human experts to efficiently validate and refine machine-generated labels.

Topic localization is related to topic segmentation and span prediction tasks such as extractive question answering, but differs in key respects. Unlike segmentation, it does not assume a partition of the text into contiguous topical blocks; and unlike QA-style span prediction, it allows multiple valid spans for the same topic and does not reduce naturally to a single answer span. These properties make both annotation and evaluation substantially more challenging, as model performance depends not only on semantic understanding but also on fine-grained and potentially subjective boundary placement.

The recent success of large language models (LLMs) raises a natural question: can LLMs perform topic localization in a manner comparable to humans? Standard evaluations typically compare predictions against a single reference annotation, implicitly assuming high and homogeneous human agreement. However, for tasks involving abstract categories and fuzzy boundaries, annotators may disagree systematically, and model scores against a single reference can be difficult to interpret without quantifying human consistency. To meaningfully assess topic localization models, evaluation should therefore be grounded in inter-annotator agreement.

To address this gap, we introduce a new human-annotated dataset for topic localization derived from Czech historical documents. The dataset consists of text excerpts from digitized archival sources, human-defined topics specified by names and descriptions, and manually annotated spans marking where these topics are expressed within the texts. Unlike existing datasets, which typically rely on fixed label sets, contiguous segmentation, or question-driven span extraction, our dataset combines open-ended topic definitions with fine-grained span localization. The key differences between our dataset and related benchmarks are summarized in Table~\ref{tab:seg-qa-datasets}. The presented dataset enables evaluation at two levels: topic presence detection and word-level span localization.

We use the proposed dataset to benchmark a diverse set of publicly available LLMs under multiple prompting configurations, as well as several fine-tuned BERT-based models, and compare their behavior to human annotators. Rather than constructing a single gold annotation, we define the human baseline using average pairwise agreement: for each annotator, agreement scores are computed against all other annotators and averaged, and these per-annotator scores are subsequently averaged across annotators. Model evaluation follows the same protocol, with each model compared against every human annotator and the resulting agreement scores averaged.

The main contributions of this paper are:
\begin{enumerate}
    \item We introduce a new human-annotated dataset for topic localization comprising 525 texts, 363 topics, and 1{,}820 annotated (text, topic) pairs. Experiments show that LLMs exhibit substantial performance variability on this dataset.
    \item We construct a large distilled development dataset and fine-tune a set of BERT-based cross-encoder models for topic localization.
    \item We quantify human agreement and benchmark LLMs and distilled BERT models under multiple configurations, analyzing performance relative to inter-annotator agreement.
\end{enumerate}

\begin{table}[ht]
    \centering
    \caption{Comparison of existing datasets across document classification, extractive question answering, topic segmentation, and named entity recognition tasks. Columns indicate whether datasets explicitly model topics, provide span-level annotations, and support an open label space defined by natural-language queries or descriptions rather than a fixed label set. Our dataset uniquely combines topic-level supervision, span annotations, and an open label space.}
    \label{tab:seg-qa-datasets}
    \resizebox{0.9\textwidth}{!}{
    \begin{tabular}{lc@{\hspace{6pt}}c@{\hspace{6pt}}c@{\hspace{6pt}}c@{\hspace{6pt}}c}
    \toprule
    Dataset & Language & Topics & \makecell{Span\\annotations} & \makecell{Open label\\space} \\
    \midrule
    20 Newsgroups \cite{LANG1995331} & en & - & \xmark & \xmark \\
    Reuters-21578 \cite{Lewis1997Reuters21578TC} & en & - & \xmark & \xmark \\
    Czech News Dataset \cite{kydlivcek2023dataset} & cs & - & \cmark & \xmark \\
    \midrule
    SQuAD \cite{squad1,squad2} & en & \xmark & \cmark & \cmark \\
    Natural Questions \cite{kwiatkowski-etal-2019-natural} & en & \xmark & \cmark & \cmark \\
    HotpotQA \cite{yang2018hotpotqa} & en & \xmark & \cmark & \cmark \\
    FEVER \cite{thorne-etal-2018-fever} & en & \xmark & \cmark & \cmark \\
    \midrule
    Choi \cite{choi-2000-advances} & en & \cmark & \cmark & \xmark \\
    Wiki-727K \cite{wiki727k} & en & \cmark & \cmark & \xmark \\
    WikiSection \cite{wikisection} & en & \cmark & \cmark & \xmark \\
    \midrule
    Universal NER \cite{mayhew2024universal} & multi &\xmark & \cmark & \xmark \\
    CNEC \cite{SevcikovaEtAl2007CNEC} & cs & \xmark & \cmark & \xmark \\
    \midrule
    CzechTopic (ours) & cs & \cmark & \cmark & \cmark \\
    \bottomrule
    \end{tabular}
    }
\end{table}

\section{Related Work}
\paragraph{Document classification.}
Traditional document classification assigns topics or other labels to entire texts and is commonly evaluated on datasets such as 20 Newsgroups \cite{LANG1995331}, AG News \cite{zhang2015character}, and Reuters \cite{Lewis1997Reuters21578TC}. Zero-shot classification generalizes this setting by allowing models to predict labels that are not fixed during training, typically by expressing candidate labels in natural language and evaluating their compatibility with the input text. In our dataset, document-level classification arises naturally: a document is considered positive for a topic if at least one span is present.

\paragraph{Topic segmentation.}
Topic segmentation partitions documents into coherent topical units by detecting topic shift boundaries in continuous text. Early approaches such as TextTiling \cite{texttiling} rely on lexical cohesion, while recent methods employ transformer-based models. Benchmark datasets typically provide contiguous segment annotations and evaluate boundary detection using metrics such as \texttt{Pk} \cite{beeferman1999statistical} and \texttt{WindowDiff} \cite{pevzner-hearst-2002-critique}. In contrast, our dataset does not assume full topic coverage or require boundary detection; instead, it focuses on localizing potentially overlapping and non-contiguous spans corresponding to human-defined topics, representing a different granularity and prediction structure from topic segmentation tasks.

\paragraph{Topic discovery.}
Topic discovery methods aim to uncover latent themes in text collections. Classical approaches such as LDA \cite{blei2003latent} model topics as distributions over words and documents, while recent methods including BERTopic \cite{grootendorst2022bertopic} leverage transformer-based embeddings and clustering, and FASTopic \cite{wu2024fastopic} further employs optimal transport for efficient topic extraction. These approaches are typically unsupervised and operate at the corpus level. In contrast, our work assumes human-defined topics specified by names and descriptions and focuses on localizing where these predefined topics occur within individual texts, rather than discovering latent topic structure.

\paragraph{Extractive question answering.}
Extractive QA is typically formulated as a span selection, as popularized by benchmarks such as SQuAD \cite{squad1,squad2} and Natural Questions \cite{kwiatkowski-etal-2019-natural}. Multi-hop datasets like HotpotQA \cite{yang2018hotpotqa} additionally provide supporting evidence annotations, linking answer prediction with rationale extraction. Unlike QA, which localizes evidence supporting a query, our work focuses on localizing spans corresponding to predefined topics, shifting the objective from pure information extraction toward semantic text analysis.

\paragraph{Span labeling with large language models.}
Span labeling is commonly solved by in-context learning with LLMs, which use one of the following output encoding paradigms: tagging, indexing, and matching \cite{semin2026strategies}. The tagging paradigm reconstructs the input text while inserting markers that denote span boundaries \cite{wang2025tagging,yan2024ltner}. The indexing paradigm requires models to predict explicit positional offsets, which has been shown to be unreliable due to limitations of LLMs in precise text positional reasoning \cite{hasanain2024large,ramponi2025finegrained}. In contrast, the matching paradigm generates only the relevant span content, and positions are recovered in post-processing via string matching \cite{semin2026strategies,kocmi2023gemba,hasanain2024large,li2023codeie}. In this paper, we focus exclusively on the tagging and matching paradigms and omit indexing baselines.

\paragraph{Fine-tuning approaches for sequence labeling.}
Beyond standard LLM prompting strategies, alternative supervised fine-tuning (SFT) methods have been proposed~\cite{duki2025supervisedincontextfinetuninggenerative,investigating2024}. In particular, \cite{investigating2024} improve sequence labeling in decoder-only models by removing the causal mask to enable bidirectional information flow. Other approaches reformulates zero-shot labeling as pairwise text--label classification: Yin et al.~\cite{yin2019benchmarking} model classification as textual entailment between input texts and label descriptions, while GLiNER~\cite{zaratiana-etal-2024-gliner} performs span extraction conditioned on natural-language labels, enabling zero-shot entity recognition.

\paragraph{Historical document retrieval.}
Recent work has also explored topic-sensitive retrieval in historical document collections. Fetch-A-Set \cite{fetchaset} is an OCR-free benchmark that uses natural-language topic queries to retrieve relevant fragments from historical document images. Unlike our work, it treats topics as retrieval queries over a visual document index rather than localizing predefined topic spans within transcribed text.

\section{Task Overview and Dataset}
\label{sec:task}
We formally define the task of \emph{topic localization} as follows.
Let a document $D$ consist of a sequence of words $W = \{w_{D1}, w_{D2}, w_{D3}, \dots, w_{DI}\}$.
Each document is associated with a topic $T = (N, S)$, where $N$ denotes the topic name and $S$ its description. The objective of topic localization is then to partition the document word sequence $W$ into two disjoint subsets $W_{\text{pos}}$ and $W_{\text{neg}}$, where $W_{\text{pos}}$ contains words that form spans corresponding to the given topic, and $W_{\text{neg}}$ contains the remaining words.

In our datasets, documents contain 130 words on average. Topic names are short phrases of at most four words, while topic descriptions contain 13 words on average.

\subsection{Data Preparation}
The data in our dataset originate from historical Czech documents, primarily books and periodicals. The dataset was compiled from digitized historical documents publicly available through the Czech Digital Library portal\footnote{\url{www.digitalniknihovna.cz}}. The data originate from 23 different Czech libraries represented within the portal. We provide metadata including the source library and URL to the corresponding public record for each text.

Each document page is available as a scanned image. 
To obtain a textual representation, we apply PERO-OCR \cite{pero1,pero2,pero3}, which produces structured OCR output including transcribed text lines.
From the OCR output, we construct continuous text, and split each page into smaller segments of 768--1024 characters.

For each text, we compute dense text embeddings using the Gemma~2 model fine-tuned for multi-task embedding~\cite{bge-m3,bge-embedding}.
These embeddings provide a semantic representation of the text content.
We then cluster the embeddings using the K-means algorithm \cite{kmeans} to group semantically similar texts. We do this to increase the difficulty of the topic localization task: rather than distinguishing between clearly unrelated themes, annotators and models must discriminate between more subtle topical differences within semantically coherent groups.

From each cluster, we select several texts closest to the cluster centroid. As a result, the dataset is organized into groups of semantically similar texts while maintaining a broad diversity of topics across clusters. Topics are subsequently defined and annotated independently for each cluster.

\subsection{Human Annotation}
To enable a reliable evaluation of the proposed task, we constructed a novel human-annotated dataset.
The annotation process involved multiple annotators and was conducted in two phases: topic definition (Phase~1) and topic localization (Phase~2), ensuring consistency with the task definition.
All annotators received detailed written guidelines specifying the definition of a topic, requirements for topic descriptions, criteria for topic granularity, and rules for span localization.
In addition, regular meetings were held throughout the annotation process to discuss ambiguous cases, refine the guidelines, and promote consistent annotation practices.

\paragraph{Phase 1: Topic Definition and Localization.}
In the first phase, an annotator is presented with five texts corresponding to a single cluster.
The annotator is instructed to propose between two and five topics that are present in the cluster. Since the texts within a cluster are semantically similar, the annotator is required to identify themes that recur across multiple texts. Each topic must satisfy the following criteria:
\begin{itemize}
    \item The topic must appear in at least two of the five texts.
    \item The topic must not be overly general (e.g., covering the whole text).
    \item The topic must not be overly specific, such that the task degenerates into for example named entity recognition.
\end{itemize}

After defining the topics, the annotator localizes each topic within the texts by marking the corresponding text spans. The topics are treated as independent labels; therefore, spans may overlap and multiple topics may be assigned to the same region of text. In this phase, each cluster is annotated by exactly one annotator. On average, the annotation of a single cluster in Phase~1 required approximately 30--45 minutes.

\paragraph{Phase 2: Topic Localization Agreement.}
In the second phase, the same five texts from a given cluster are presented again. In contrast to Phase~1, the annotator is provided with the set of topics already defined in Phase~1. The task in this phase is solely to localize each predefined topic within the texts by marking the corresponding spans.

Each cluster is annotated by multiple annotators to enable the measurement of inter-annotator agreement for the localization task while keeping the topic definitions fixed. Annotators in Phase~2 do not have access to the span annotations produced by other annotators, ensuring independent judgments. The annotation of a single cluster in this phase required approximately 15--25 minutes on average.

This two-phase annotation procedure ensures that the dataset reflects the textual definitions of topics rather than relying on interpretations held by the topic author. At the same time, this enables analysis of how well the written topic definitions capture the original intent of the topic author, allowing the evaluation of both the clarity of topic formulations and the extent to which independent annotators interpret and localize topics in a consistent manner.

\subsection{Resulting Dataset}
\begin{figure}[ht]
    \centering
    \includegraphics[width=\linewidth]{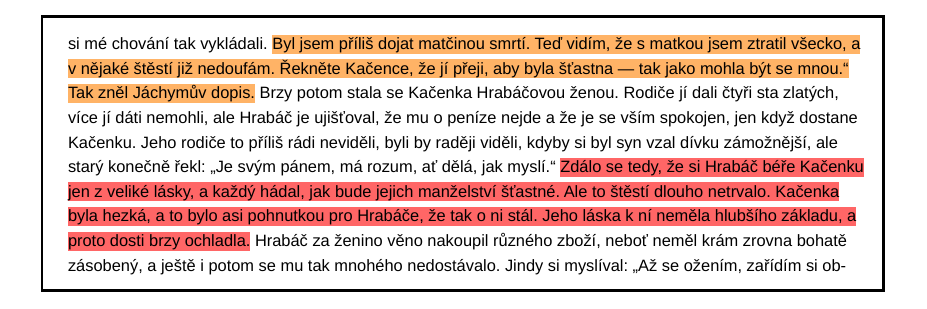}
    \caption{Example from our topic localization dataset showing a text annotated with spans belonging to two distinct topics.}
    \label{fig:annotated-example}
\end{figure}

\begin{figure}[ht]
    \centering
    \begin{subfigure}{0.32\textwidth}
    \includegraphics[width=\linewidth]{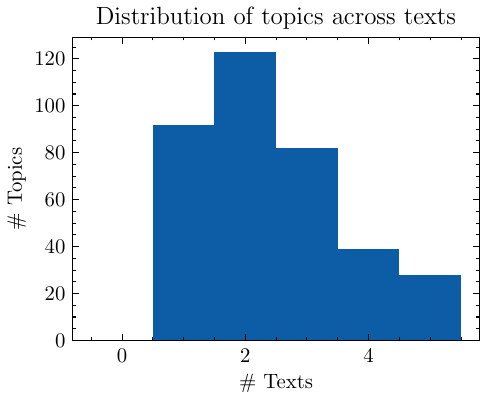}
    \caption{Phase 1}
    \label{fig:topic-distibution:subfig1}
    \end{subfigure}
    \hfill
    \begin{subfigure}{0.32\textwidth}
    \includegraphics[width=\linewidth]{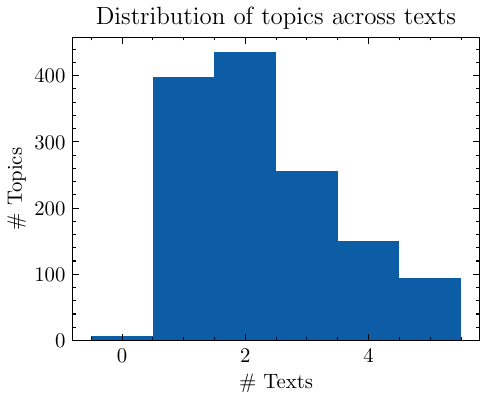}
    \caption{Phase 2}
    \label{fig:topic-distribution:subfig2}
    \end{subfigure}
    \hfill
    \begin{subfigure}{0.32\textwidth}
    \includegraphics[width=\linewidth]{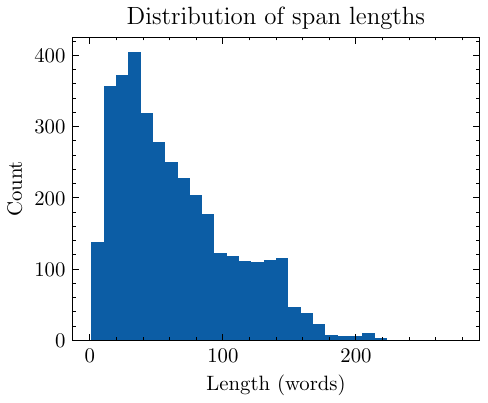}
    \caption{Phase 2}
    \label{fig:topic-distribution:subfig3}
    \end{subfigure}
    \caption{Distribution statistics of our topic localization dataset. Distribution of the number of texts in which a topic appears in Phase~1 (a) and Phase~2 (b). The Phase~2 distribution is aggregated across all annotators. Note that some topics are not annotated by certain annotators, resulting in zero occurrences for those annotator--topic pairs. (c) Distribution of annotated span lengths measured in words for Phase~2 annotations.}
    \label{fig:topic-distribution}
\end{figure}

The resulting dataset comprises 105 annotated text clusters, each containing five texts (525 texts in total).
In total, 363 topics were annotated, yielding 1,820 annotated (text, topic) pairs.
Figure~\ref{fig:annotated-example} presents an example text from the dataset annotated with two distinct topics.
Figure~\ref{fig:topic-distribution} illustrates the distribution of topic occurrences across texts as well as the distribution of annotated span lengths.

\subsection{LLM Distillation}
\label{sec:distilled-dataset}
The human annotation process described above is time-consuming and difficult to scale to larger datasets.
To enable model development and training at scale, we constructed a separate development dataset using LLM distillation.

Specifically, the \texttt{gpt-5-mini} model was employed to generate annotations in a manner analogous to the human annotation procedure.
The generation process was also performed in two stages, mirroring the structure of the human annotation pipeline.

\paragraph{Phase 1: Topic Generation.}
In the first phase, the model was provided with 15 texts corresponding to a single cluster.
Based on the joint content of these texts, the model was instructed to generate between 5 and 20 candidate topics that recur across the cluster.
The prompt design followed the same principles as the human guidelines, encouraging semantically meaningful topics that are neither overly general nor overly specific.

\paragraph{Phase 2: Topic Localization.}
In the second phase, the model was presented with a single text together with the complete list of topics generated for the corresponding cluster.
For each topic, the model was required to localize relevant text spans within the text using an matching approach.

This two-phase distillation procedure enabled the creation of a substantially larger annotated development dataset while maintaining structural consistency with the human annotation protocol.
Although automatically generated, the distilled annotations preserve the key characteristics of the task formulation and provide a practical foundation for training and experimentation.
The resulting dataset comprises 1,550 clusters, each containing ten texts (15,550 texts in total). In total, 19,107 topics were annotated, yielding 187,773 annotated (text, topic) pairs.

\section{Experiments}
This section describes the evaluation protocol and experimental setup used to estimate human agreement and benchmark model performance.

\subsection{Evaluation}
We evaluate the task at two levels: text-level and word-level.
At the text level, the task is to determine whether a given topic is present in a text.
We report precision, recall, and F1 score. The F1 score is accompanied by a 95\% confidence interval computed via bootstrap resampling over topics.

At the word level, the task is to identify which words correspond to a given topic, as defined in Section \ref{sec:task}. We report precision, recall, F1, and Intersection over Union (IoU), with confidence intervals provided for F1 and IoU.
In addition, we define localization success, where a topic is considered successfully localized if the IoU exceeds a predefined threshold.

All metrics are macro-averaged over topics.
The human baseline is computed as the average pairwise performance among annotators.
For each annotator, scores are calculated by comparing their annotations against those of all remaining annotators and averaging the results.
The final human baseline is obtained by averaging these per-annotator scores across all annotators.

\subsection{BERT Fine-Tuning}
\begin{figure}[ht]
    \centering
    \includegraphics[width=\textwidth]{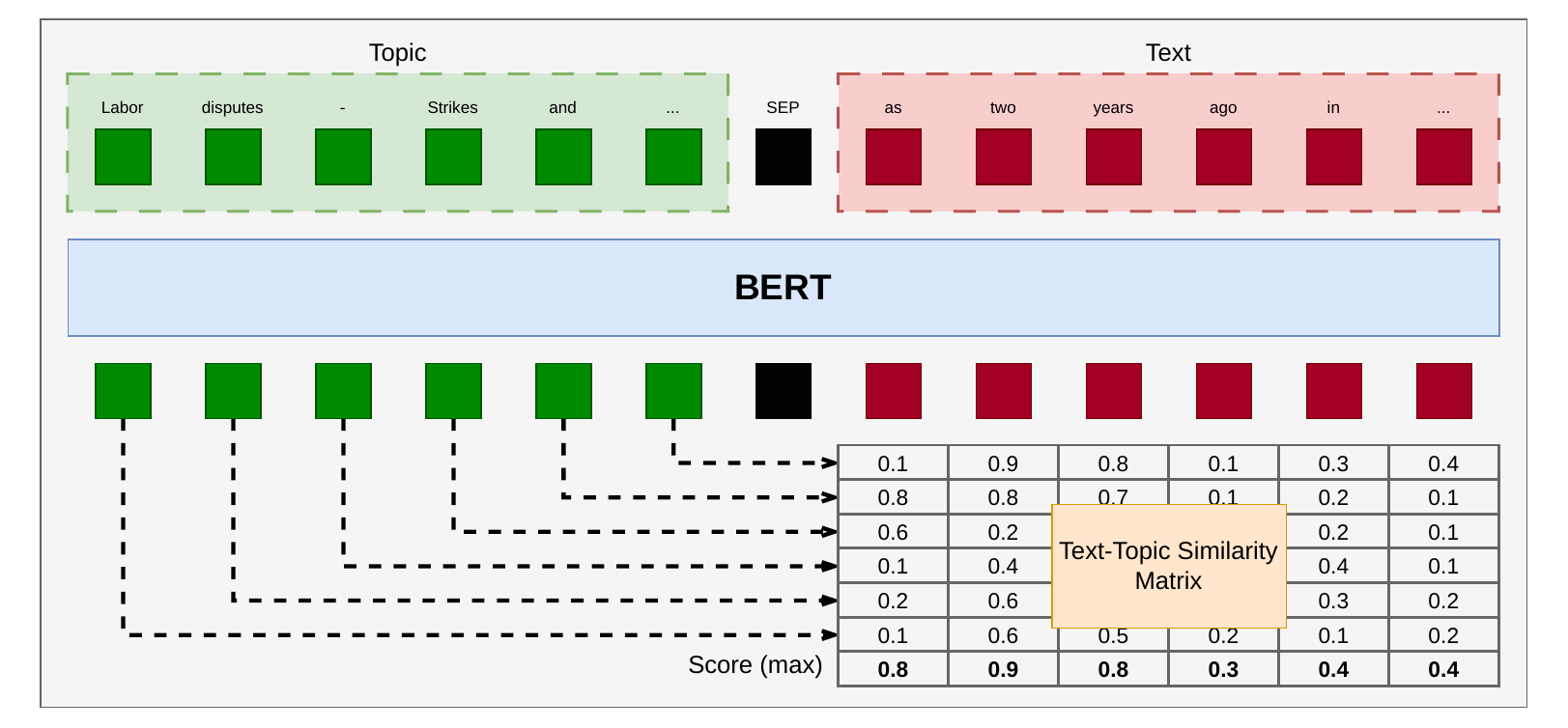}
    \caption{Overview of the cross-encoder architecture used for topic localization. The topic and text are jointly encoded by BERT. A similarity matrix is computed between topic tokens and text tokens, and each text token receives a score given by the maximum similarity across all topic tokens.}
    \label{fig:bert}
\end{figure}

We draw inspiration from GLiNER \cite{zaratiana-etal-2024-gliner}, which demonstrates zero-shot labeling using a cross-encoder formulation. We fine-tune a set of BERT-based models on the development dataset described in Section~\ref{sec:distilled-dataset}. The dataset is split into training and validation subsets using an 80:20 partition over clusters, ensuring that both topics and texts remain disjoint between splits. Each model is trained as a cross-encoder, jointly encoding the topic description and the text as input, as illustrated in Figure~\ref{fig:bert}.

Given the joint encoding, we compute a similarity matrix between topic tokens and text tokens.
A score is then assigned to each text token by taking the maximum similarity across all topic tokens.
Models are trained using an effective batch size of 64, the AdamW optimizer, and a learning rate of $3\times10^{-5}$.
For each model, the final classification threshold is selected based on the word-level F1 score measured on the validation split.

\subsection{Large Language Models}
We evaluated several open and commercial large language models.
Two strategies for span prediction were explored: tagging and matching.
Experiments were conducted in both zero-shot and few-shot settings; in the latter, models were provided with two annotated examples illustrating span localization.
Finally, we examined the effect of prompt language by comparing prompts written in Czech and English. The prompts and model-specific settings used in the inference are included in the provided GitHub repository.

\subsection{Results}
\begin{figure}[ht]
    \noindent
    \hspace{2.8pt}
    \begin{subfigure}{0.49\textwidth}
        \includegraphics[scale=0.717]{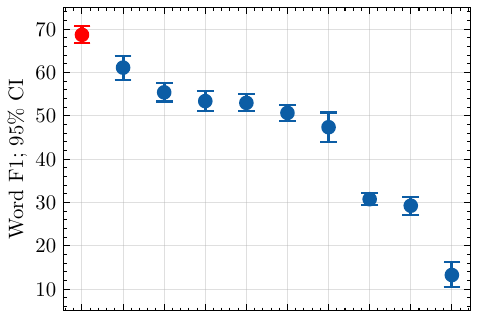}
    \end{subfigure}
    \hspace{22.2pt}
    \begin{subfigure}{0.49\textwidth}
        \includegraphics[scale=0.717]{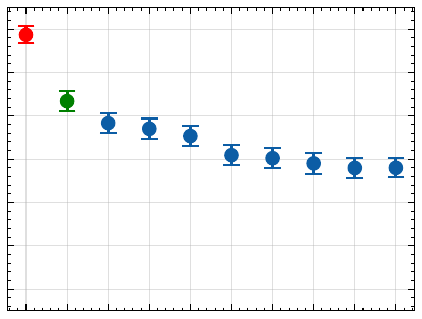}
    \end{subfigure}
    \begin{subfigure}{0.49\textwidth}
        \includegraphics[width=\linewidth]{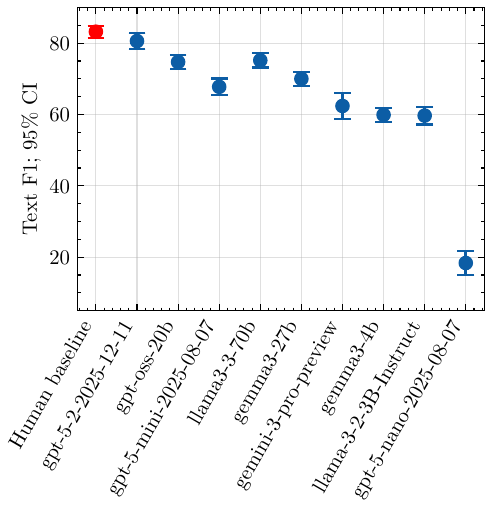}
        \caption{Large language models}
    \end{subfigure}
    \hfill
    \begin{subfigure}{0.49\textwidth}
        \includegraphics[width=\linewidth]{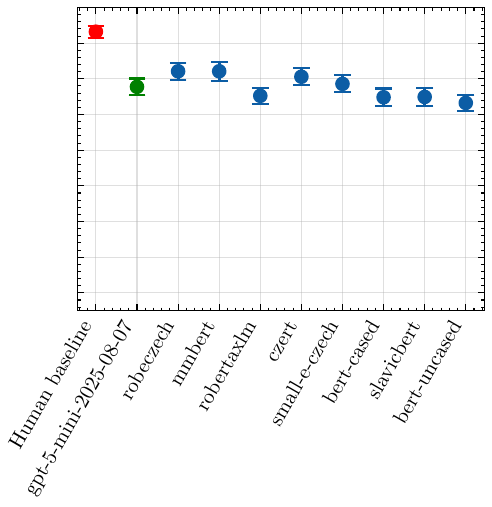}
        \caption{BERT-based models}
    \end{subfigure}
    \caption{Comparison of model performance. The top row shows word-level F1 scores and the bottom row text-level F1 scores, each with 95\% confidence intervals. (a) Large language models and (b) BERT-based models. The human baseline is highlighted in red. In (b), the teacher model is highlighted in green.}
    \label{fig:model-comparisons}
\end{figure}

\begin{figure}[ht]
    \centering
    \begin{subfigure}{0.49\textwidth}
        \includegraphics[width=\linewidth]{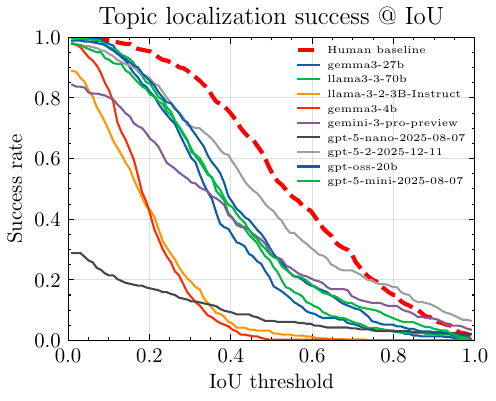}
        \caption{Large language models}
    \end{subfigure}
    \hfill
    \begin{subfigure}{0.49\textwidth}
        \includegraphics[width=\linewidth]{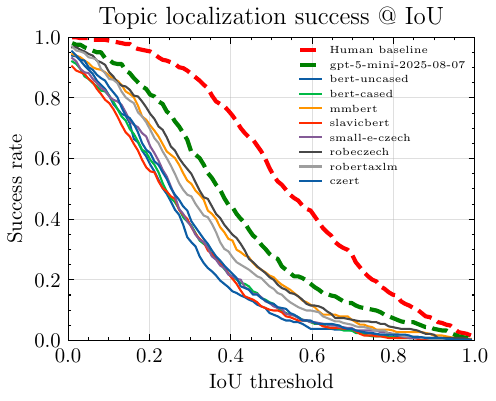}
        \caption{BERT-based models}
    \end{subfigure}
    \caption{Comparison of model performance with respect to topic localization success across IoU thresholds. (a) Large language models and (b) BERT-based models. Curves show the proportion of topics successfully localized at a given IoU threshold. The human baseline is highlighted in red, while in (b) the teacher model is highlighted in green.}
    \label{fig:success}
\end{figure}

Table~\ref{tab:human_model_results} presents the evaluation results of all human annotators, BERT-based models, and large language models on both text-level and word-level tasks. Figures~\ref{fig:model-comparisons} and~\ref{fig:success} provide a visual comparison of model performance against the human baseline and illustrate topic localization success across IoU thresholds, respectively.
\begin{table}[ht]
    \caption{Text-level and word-level performance of human annotators and evaluated models on the topic localization task. Text-level metrics measure topic presence detection, while word-level metrics evaluate word-level localization quality. Results are reported as macro averages over topics. Confidence intervals (shown in brackets) denote 95\% bootstrap intervals computed over topics for F1 and IoU. Models are grouped by category: human annotators, BERT models, and large language models. Each group is ordered by word-level F1.}
    \label{tab:human_model_results}
    
    \resizebox{\textwidth}{!}{
    \begin{tabular}{lc@{\hspace{4pt}}c@{\hspace{4pt}}c@{\hspace{4pt}}|@{\hspace{4pt}}c@{\hspace{4pt}}c@{\hspace{4pt}}c@{\hspace{4pt}}c}
    \toprule
     & \multicolumn{3}{c}{Text-level} & \multicolumn{4}{c}{Word-level} \\
     & P & R & F1 & P & R & F1 & IoU \\
    \midrule
    Human baseline & $85.7$ & $86.7$ & $83.2_{[81.5, 84.9]}$ & $72.8$ & $72.8$ & $68.7_{[66.7, 70.6]}$ & $57.2_{[55.1, 59.3]}$ \\
    \midrule
    Annotator 6 & $89.4$ & $88.3$ & $86.4_{[84.6, 88.2]}$ & $79.5$ & $72.8$ & $72.1_{[69.7, 74.4]}$ & $61.0_{[58.2, 63.7]}$ \\
    Annotator 3 & $88.5$ & $84.7$ & $83.9_{[81.9, 85.7]}$ & $73.4$ & $73.2$ & $69.5_{[67.3, 71.7]}$ & $58.2_{[55.7, 60.7]}$ \\
    Annotator 4 & $84.3$ & $87.1$ & $82.8_{[80.3, 85.1]}$ & $70.3$ & $75.7$ & $68.8_{[66.2, 71.4]}$ & $57.5_{[54.6, 60.4]}$ \\
    Annotator 7 & $83.8$ & $88.3$ & $83.1_{[81.1, 85.1]}$ & $72.3$ & $73.5$ & $68.5_{[66.1, 70.8]}$ & $56.7_{[54.1, 59.3]}$ \\
    Annotator 5 & $82.7$ & $84.8$ & $80.3_{[77.7, 82.9]}$ & $71.0$ & $71.1$ & $66.4_{[63.7, 69.0]}$ & $55.0_{[52.2, 57.9]}$ \\
    \midrule
    robeczech \cite{robeczech} & $67.4$ & $91.2$ & $72.1_{[69.7, 74.5]}$ & $60.7$ & $51.7$ & $48.3_{[46.0, 50.6]}$ & $35.5_{[33.3, 37.7]}$ \\
    mmbert \cite{mmbert}& $73.5$ & $84.6$ & $72.1_{[69.5, 74.7]}$ & $67.2$ & $45.4$ & $47.0_{[44.6, 49.4]}$ & $34.7_{[32.5, 36.9]}$ \\
    robertaxlm \cite{conneau-etal-2020-unsupervised}& $56.2$ & $94.2$ & $65.2_{[63.0, 67.4]}$ & $65.6$ & $41.7$ & $45.3_{[43.1, 47.6]}$ & $32.7_{[30.7, 34.7]}$ \\
    czert \cite{czert} & $68.7$ & $84.2$ & $70.6_{[68.2, 73.0]}$ & $61.1$ & $38.4$ & $40.9_{[38.6, 43.2]}$ & $28.8_{[26.9, 30.9]}$ \\
    small-e-czech \cite{small-e-czech} & $68.1$ & $82.8$ & $68.6_{[66.2, 71.0]}$ & $58.0$ & $41.8$ & $40.2_{[37.9, 42.5]}$ & $28.4_{[26.4, 30.4]}$ \\
    bert-cased \cite{bert} & $60.1$ & $86.0$ & $64.8_{[62.3, 67.3]}$ & $52.1$ & $42.8$ & $39.0_{[36.7, 41.4]}$ & $27.4_{[25.4, 29.4]}$ \\
    slavicbert \cite{slavicbert} & $60.8$ & $85.0$ & $64.9_{[62.4, 67.5]}$ & $53.5$ & $40.6$ & $38.0_{[35.6, 40.3]}$ & $26.5_{[24.5, 28.5]}$ \\
    bert-uncased \cite{bert} & $55.6$ & $90.2$ & $63.3_{[61.0, 65.5]}$ & $49.8$ & $42.1$ & $38.0_{[35.8, 40.2]}$ & $26.2_{[24.4, 28.1]}$ \\
    \midrule
    gpt-5-2-2025-12-11 \cite{openai_gpt5_2_2025}& $76.1$ & $94.2$ & $80.6_{[78.3, 82.8]}$ & $63.4$ & $71.4$ & $61.1_{[58.4, 63.9]}$ & $48.7_{[45.8, 51.7]}$ \\
    gpt-oss-20b \cite{openai2025gptoss120bgptoss20bmodel} & $69.2$ & $91.3$ & $74.7_{[72.7, 76.7]}$ & $57.7$ & $64.9$ & $55.4_{[53.3, 57.5]}$ & $42.1_{[40.0, 44.2]}$ \\
    gpt-5-mini-2025-08-07 \cite{openai_gpt5_2_2025} & $64.9$ & $82.4$ & $67.8_{[65.4, 70.1]}$ & $50.6$ & $72.5$ & $53.4_{[51.0, 55.7]}$ & $40.9_{[38.6, 43.3]}$ \\
    llama3-3-70b \cite{grattafiori2024llama} & $67.2$ & $96.9$ & $75.2_{[73.2, 77.2]}$ & $56.1$ & $61.8$ & $53.0_{[51.0, 54.9]}$ & $39.2_{[37.2, 41.2]}$ \\
    gemma3-27b \cite{gemmateam2025gemma3technicalreport} & $59.9$ & $95.9$ & $70.0_{[68.1, 72.0]}$ & $50.8$ & $62.3$ & $50.7_{[48.7, 52.5]}$ & $36.8_{[34.9, 38.7]}$ \\
    gemini-3-pro-preview \cite{team2023gemini} & $72.2$ & $65.1$ & $62.4_{[58.7, 66.0]}$ & $63.9$ & $45.7$ & $47.4_{[43.9, 50.8]}$ & $36.6_{[33.4, 39.8]}$ \\
    gemma3-4b \cite{gemmateam2025gemma3technicalreport} & $47.4$ & $95.9$ & $59.9_{[57.9, 61.9]}$ & $32.8$ & $36.7$ & $30.7_{[29.3, 32.2]}$ & $19.3_{[18.2, 20.4]}$ \\
    llama-3-2-3B-Instruct \cite{grattafiori2024llama} & $49.4$ & $89.5$ & $59.7_{[57.2, 62.2]}$ & $27.8$ & $40.1$ & $29.2_{[27.2, 31.3]}$ & $18.6_{[17.1, 20.1]}$ \\
    gpt-5-nano-2025-08-07 \cite{openai_gpt5_2_2025} & $64.6$ & $16.3$ & $18.4_{[15.0, 21.9]}$ & $54.2$ & $11.9$ & $13.2_{[10.5, 16.2]}$ & $10.3_{[7.9, 12.8]}$ \\
    \bottomrule
    \end{tabular}
    }
\end{table}

\paragraph{Human annotators show a high level of agreement.}
Inter-annotator agreement measured using Krippendorff’s $\alpha$ reached 0.616 (micro) and 0.592 when macro-averaged over topics. Bootstrapping over topics yielded 95\% confidence intervals of [0.591, 0.640] and [0.567, 0.616], respectively. Individual annotators achieve word-level F1 scores between 66.4 and 72.1, confirming consistent localization behavior across annotators.

\paragraph{Large language models exhibit substantial performance variability.}
Performance varies considerably across models, with a 47.9 percentage-point gap in word-level F1 between the best and worst systems (61.1 vs.\ 13.2). While stronger models such as GPT-5-2 approach human-level performance in topic detection, smaller or less capable models degrade substantially, demonstrating that the benchmark poses a non-trivial challenge.

\paragraph{Large language models underperform human annotators.}
Even the strongest LLM achieves a word-level F1 of 61.1 and IoU of 48.7, remaining notably below the human baseline (68.7 F1 and 57.2 IoU, $p < 0.001$). This gap indicates that precise topic localization remains challenging for current generative models despite strong document-level understanding. Moreover, the best model remains significantly below human agreement even at the text level (80.6 vs.\ 83.2 F1, $p = 0.026$).

\paragraph{BERT-based models achieve competitive performance relative to their teacher model and outperform several LLMs.}
Fine-tuned token embedding models reach word-level F1 scores up to $48.3$ (robeczech; 95\% CI: $[46.0, 50.6]$) and text-level F1 scores around $72.1$ (95\% CI: $[69.7, 74.5]$), surpassing multiple LLM configurations, including smaller Gemma and Llama variants. Although still below human performance ($p < 0.001$), these results show that specialized cross-encoder architectures trained directly for localization can compete with, and in some cases exceed, general-purpose large language models.

\paragraph{Human annotators are more consistent with one another than with the topic author.}
We tested the hypothesis that annotators performing localization under fixed topic definitions exhibit higher mutual agreement than agreement with the original topic creator. Agreement among Phase~2 annotators was significantly higher than agreement between Phase~1 and Phase~2 annotations ($\Delta$F1 = 0.024, 95\% CI [0.009, 0.041], $p = 0.0013$). This result indicates a systematic, though moderate, divergence between the original topic formulation and subsequent localization judgments. In particular, it suggests that part of the topic interpretation remains implicit to the author and is not fully captured by the written topic description.

\paragraph{Ablation on the LLM setting.}
Table~\ref{tab:llm-factors} presents an ablation study analyzing the impact of configuration choices in the LLM-based pipeline.
The choice of span extraction strategy has the largest effect on performance: the matching approach improves word-level F1 by $0.104$ compared to tagging, with a narrow confidence interval indicating a consistent and substantial gain ($p < 0.001$).
In contrast, moving from zero-shot to few-shot prompting yields only a modest improvement of $0.010$ F1, suggesting limited benefit from additional examples for this task ($p < 0.001$).
Finally, prompt language has no measurable impact on performance, as the difference between Czech and English prompts is statistically negligible ($p = 0.962$).
\begin{table}[ht]
    \centering
    \caption{Ablation study of LLM configuration factors. The table reports changes in word-level F1 ($\Delta$F1), together with 95\% confidence intervals. The span extraction strategy has the largest impact, while few-shot prompting yields only marginal improvements and prompt language shows no statistically significant effect.}
    \label{tab:llm-factors}
    \begin{tabular}{lcc}
    \toprule
    Factor & $\Delta$F1 & CI \\
    \midrule
    Matching (x Tagging) & 0.104 & [0.097; 0.110] \\
    Zero-shot (x Few-shot) & 0.010 & [0.007; 0.014] \\
    Czech prompt (x English prompt) & 0.000 & [-0.003; 0.003] \\
    \bottomrule
    \end{tabular}
\end{table}

\section{Conclusions}
We presented a new human-annotated benchmark for \emph{topic localization}, a task that requires identifying spans of text expressing a given topic defined by a name and description. The dataset is derived from historical Czech documents and contains 525 texts annotated with 363 human-defined topics, yielding 1,820 annotated (text, topic) pairs. The dataset is publicly available online\footnote{The dataset is available at: \url{https://zenodo.org/records/18877204} \\ The implementation is available at: \url{https://github.com/dcgm/czechtopic}.}.

To enable scalable experimentation, we constructed a large distilled development dataset generated using the \texttt{gpt-5-mini} model and used it to fine-tune several BERT-based models.

We evaluated both large language models and fine-tuned encoder architectures against human annotators, framing evaluation relative to measured human agreement rather than a single gold annotation.

Our experiments show that topic localization remains challenging: while the strongest LLM approaches human-level performance, substantial variability exists across models, and most systems remain below human consistency, particularly at the span level. Fine-tuned BERT-based models achieve competitive performance relative to their teacher model and even surpass it on text-level evaluation, demonstrating that specialized architectures trained for localization remain strong baselines.

We hope this benchmark establishes topic localization as a distinct evaluation setting and encourages future work on structured topic understanding, improved span localization methods, and evaluation paradigms grounded in human agreement.

\begin{credits}
\subsubsection{\ackname} This work has been supported by the Ministry of Culture of the Czech Republic under the NAKI III research project semANT – Semantic Document Exploration (DH23P03OVV060).


\subsubsection{\discintname}
The authors have no competing interests to declare that are relevant to the content of this article.
\end{credits}
%
%
%
%
\bibliographystyle{splncs04}
\bibliography{refs}

\end{document}